\documentclass[letterpaper, 10 pt, journal, twoside]{IEEEtran}
\ifCLASSINFOpdf
\else
\fi
\usepackage{multirow}
\usepackage{lscape} 
\usepackage{tabularx} 
\usepackage{graphicx} 
\usepackage{comment}
\usepackage{color}
\usepackage{algorithm}
\usepackage[noend]{algpseudocode} 
\usepackage{amssymb,amsmath}
\newtheorem{theorem}{Theorem}

\newtheorem{definition}{Definition}

\newcommand{\blue}{}

\newcommand{\gv}{\mathbf{g}}
\newcommand{\vv}{\mathbf{v}}
\newcommand{\fv}{\mathbf{f}}
\newcommand{\hv}{\mathbf{h}}

\usepackage{etoolbox}
\makeatletter
\patchcmd{\@makecaption}
  {\scshape}
  {}
  {}
  {}
\patchcmd{\@makecaption}
{\\}
{.\ }
{}
{}
\makeatother

\begin{document}
	
\title{Multi-Objective Path-Based D* Lite}

\author{Zhongqiang Ren$^{1}$, Sivakumar Rathinam$^{2}$, Maxim Likhachev$^{1}$ and Howie Choset$^{1}$
\thanks{Zhongqiang Ren, Maxim Likhachev and Howie Choset are with Carnegie Mellon University, 5000 Forbes Ave., Pittsburgh, PA 15213, USA. (email: zhongqir@andrew.cmu.edu; maxim@cs.cmu.edu; choset@andrew.cmu.edu).}%
\thanks{Sivakumar Rathinam is with Texas A\&M University, College Station, TX 77843-3123. (email: srathinam@tamu.edu).}
\thanks{Digital Object Identifier (DOI): see top of this page.}
}

	\maketitle
	
    \graphicspath{{./figures/}}

	\begin{abstract}
		Incremental graph search algorithms such as D* Lite reuse previous, and perhaps partial, searches to expedite subsequent path planning tasks. In this article, we are interested in developing incremental graph search algorithms for path finding problems to simultaneously optimize multiple objectives such as travel risk, arrival time, etc. This is challenging because in a multi-objective setting, the number of ``Pareto-optimal'' solutions can grow exponentially with respect to the size of the graph. This article presents a new multi-objective incremental search algorithm called Multi-Objective Path-Based D* Lite (MOPBD*) which leverages a path-based expansion strategy to prune dominated solutions. Additionally, we introduce a sub-optimal variant of MOPBD* to improve search efficiency while approximating the Pareto-optimal front. We numerically evaluate the performance of MOPBD* and its variants in various maps with two and three objectives. Results show that our approach is more efficient than search from scratch, and runs up to an order of magnitude faster than the existing incremental method for multi-objective path planning.
	\end{abstract}
	
\begin{IEEEkeywords}
	Motion and Path Planning, Planning under Uncertainty
\end{IEEEkeywords}
	
	\section{Introduction}\label{sec:intro}
	

\IEEEPARstart{T}{he} Shortest Path Problem (SPP), which aims {\blue of finding} a minimum-cost path between two nodes in a graph, is a problem of fundamental importance with numerous applications in robotics and logistics \cite{urmson2008autonomous,wurman2008coordinating}. There are several algorithms \cite{astar,deo1984shortest} that can solve SPP to optimality.
Incremental search algorithms, such as LPA*~\cite{koenig2004lifelong}, D* Lite~\cite{koenig2005fast} etc., generalize these planners to a dynamic setting that allows for cost changes in the edges of the graph. 
When edge costs change, incremental search algorithms aim to reuse previous searches to speed up subsequent planning tasks. Incremental search is very useful in robotic applications which include navigation in an unknown terrain~\cite{urmson2008autonomous}.

\begin{figure}[htbp]
	\centering
	\includegraphics[width=0.9\linewidth]{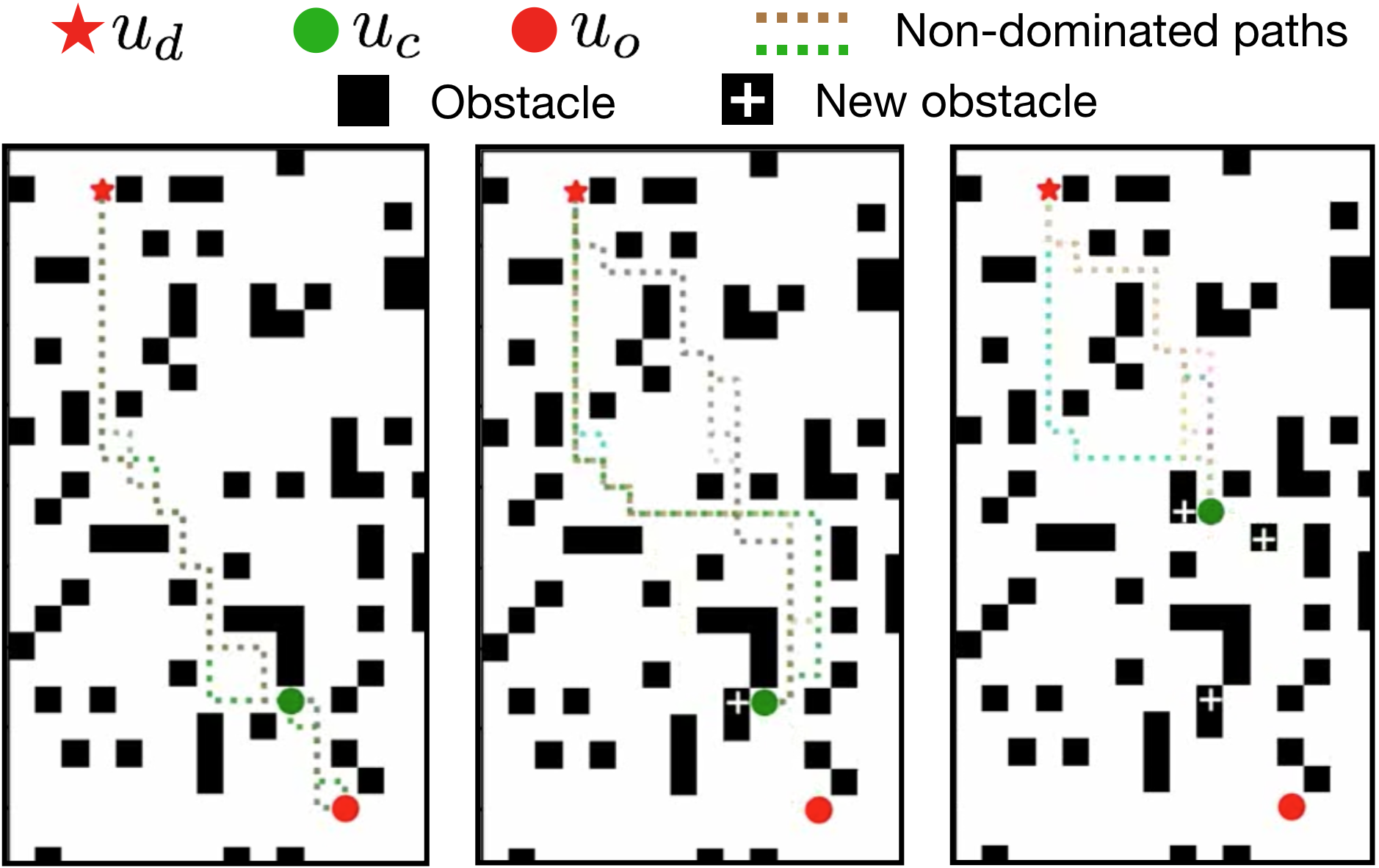}
	\caption{An illustration of both the problem and the simulator used in the test. On the left, initial planning is finished and the robot is following a selected path (Step-1 and Step-2 in simulation) towards its destination $u_d$. In the middle, a new obstacle is detected in front of the robot and paths are replanned (Step-3 and Step-4 in simulation). On the right, the simulator keeps running after detecting three new obstacles.}
	\label{fig:test_sim}
\end{figure}

One can envision applications such as hazardous material transportation~\cite{bronfman2015maximin}, robot routing in urban waterways~\cite{shan2020receding}, where path planning may involve minimizing multiple (conflicting) objectives such as travel risk, fuel usage, arrival time, to name a few.
It may not be possible to convert these objectives into a single, weighted objective because the choice of weights is difficult to obtain~\cite{roijers2013survey}.
This has led researchers to address the multi-objective shortest path problem (MO-SPP)~\cite{loui1983optimal,moastar}. MO-SPP generalizes the conventional SPP by associating each edge in the given graph with a cost vector where each component of the vector corresponds to an objective to be minimized.

In the presence of multiple conflicting objectives, in general, there may not be a single path that optimizes all the objectives.
Therefore, the goal of MO-SPP is to find a Pareto-optimal set (of feasible paths), whose cost vectors form the so-called Pareto-optimal front.
A path is Pareto-optimal (or non-dominated) if no objective can be improved without deteriorating at least one of the other objectives.
MO-SPP is NP-hard, even with two objectives~\cite{hansen1980bicriterion,ehrgott2005multicriteria}, as the size of the Pareto-optimal front can grow exponentially with respect to the number of nodes in the graph.
To solve MO-SPP, there are several A*-like planners~\cite{moastar,mandow2008multiobjective,ulloa2020simple,goldin2021approximate} which compute the exact or an approximated Pareto-optimal front.

In this work, we consider the dynamic version of the MO-SPP where the costs of the edges can change. After an event when some edge costs change, we aim to develop an incremental search algorithm that reuses previous searches to speed up similar planning tasks. Incremental search is important since a naive approach that computes Pareto-optimal front from scratch after every event can be computationally expensive for MO-SPP.
To our knowledge, the only existing work that considers a similar problem is MOD*~\cite{oral2015mod}, which combines D* Lite and MOA*~\cite{moastar} to reuse previous searches.
However, MOA* has been shown~\cite{mandow2008multiobjective} to be inefficient due to its {\it node-based} expansion during the search and is outperformed by NAMOA*~\cite{mandow2008multiobjective} which employs {\it path-based} expansion. (See Sec.~\ref{sec:preliminary} for details).

This work aims to leverage both D* Lite and the path-based expansion in NAMOA* to create a novel incremental multi-objective planner to plan paths in dynamic environments.
Fusing D* Lite and NAMOA* is challenging. D* Lite defines local consistency between adjacent nodes and keeps expanding nodes that are locally inconsistent until an optimal path is found; it's non-trivial to fuse local consistency between \emph{nodes} with a \emph{path-based} expansion strategy in a straightforward way. 

To achieve this goal, we propose a new type of local consistency in multi-objective settings that is suitable for path-based methods.
We then develop a new algorithm named Multi-Objective Path-Based D* Lite (MOPBD*). We analyze and show that MOPBD* is able to compute all Pareto-optimal solutions in a dynamic graph. In addition, we also develop a sub-optimal variant of MOPBD* called MOPBD*-$\epsilon$, which leverages $\epsilon$-dominance~\cite{perny2008near} to efficiently approximate the Pareto-optimal front.
To verify the proposed algorithms, we run extensive numerical simulations in several dynamic graphs with two and three objectives, which show MOPBD* and its variant are more efficient in comparison with running NAMOA* from scratch (baseline 1) and the existing node-based incremental method MOD* (baseline 2).

	\section{Related Work}\label{sec:related}
	
Incremental search algorithms such as D*~\cite{stentz1995focussed}, LPA*~\cite{koenig2004lifelong}, D* Lite~\cite{koenig2005fast}, etc~\cite{aine2013anytime}, reuse previous searches (by storing partial solution paths in a search tree) to speed up the current search without losing optimality guarantees.
However, all these algorithms optimize a single-objective: minimizing the sum of edge cost values along the planned path.

In the case of multiple objectives, developing efficient algorithms to compute the exact Pareto-optimal set or its approximation for MO-SPP has a long history~\cite{loui1983optimal} and remains an active research topic~\cite{moastar,mandow2008multiobjective,ulloa2020simple,goldin2021approximate}.
Seminal works MOA*~\cite{moastar} and NAMOA*~\cite{mandow2008multiobjective} both extend A* to handle multiple objectives but with different strategies. 
MOA* uses a node-based selection and expansion strategy while NAMOA* adopts a path-based one and outperforms MOA* in general~\cite{mandow2008multiobjective}.

This work focuses on incremental search algorithms for MO-SPP.
To our limited knowledge, the only work in this area is MOD*~\cite{oral2015mod}, which is an incremental node-based multi-objective search algorithm combining the best of MOA* and D* Lite. 
Since it is known that NAMOA* outperforms MOA* in many instances~\cite{mandow2008multiobjective}, we expect an incremental search algorithm based on NAMOA* can also outperform search algorithms (such as MOD*) based on MOA*.
Our contribution in this work is a novel algorithm named Multi-Objective Path-Based D* Lite (MOPBD*), which leverages both path-based expansion in NAMOA* and incremental search in D* Lite.
We compared our MOPBD* with both MOD*, a node-based incremental method (baseline 1), and running NAMOA* to search from scratch each time (baseline 2).
The numerical results show that MOPBD* outperforms both baselines.

	
	\section{Problem Description}\label{sec:problem}
	
Let $\mathcal{G} = (\mathcal{U}, \mathcal{E})$ denote {\blue an un-directed graph} representing the workspace of the robot, where the node set $\mathcal{U}$ denotes the set of possible locations for the robot and the {\blue edge set} $\mathcal{E}=\mathcal{U}\times\mathcal{U}$ denotes the set of actions that move the robot between any two nodes in $\mathcal{U}$. 
In addition, {\blue let $ngh(u)$ denote the set of neighbors ($i.e.$ adjacent nodes)} of a node $u \in \mathcal{U}$.
We use $u_o,u_d\in \mathcal{U}$ to denote the initial and destination node of the robot respectively, and let $u_c$ represent the current node of the robot during the navigation. Note that when the navigation starts, $u_c=u_o$.
For any two distinct nodes $w,u \in \mathcal{U}$, the edge between $w$ and $u$ is denoted as $(w,u)\in \mathcal{E}$. The cost of an edge $e\in\mathcal{E}$ is a non-negative cost vector $\vec{c}(w,u) \in (\mathbb{R}^{+})^{M}$, {\blue where $\mathbb{R}^{+}$ denotes the set of non-negative real numbers and $M$ is a positive integer.}
Here, each component of the cost vector corresponds to an objective to be minimized.

In this work, let $\pi(u_1,u_\ell)$ represent a path connecting $u_1,u_\ell \in \mathcal{U}$ via a sequence of nodes $(u_1,u_2,\dots,u_\ell)$ in $\mathcal{G}$, where $u_k$ and $u_{k+1}$ are connected by an edge $(u_k,u_{k+1}) \in \mathcal{E}$, for $k=1,2,\dots,\ell-1$.
Let $\vec{g}(\pi(u_1,u_\ell))$ denote the cost vector corresponding to the path, which is the sum of cost vectors of all edges present in the path, $i.e.$ $\vec{g}(\pi(u_1,u_\ell)) = \Sigma_{k=1,2,\dots,\ell-1} \vec{c}(u_k,u_{k+1})$.
To compare any two paths, we compare the cost vector associated with them using the dominance relationship~\cite{ehrgott2005multicriteria}:
\begin{definition}[Dominance]
	Given two vectors $a$ and $b$ of length $M$, $a$ dominates $b$ (referred as $a \succeq b$) if and only if $a(m) \leq b(m)$ $\forall m \in \{1,2,\dots,M\}$, and there exists $m\in \{1,2,\dots,M\}$ such that $a(m) < b(m)$.
\end{definition}
If $a$ does not dominate $b$, this non-dominance is denoted as $a \nsucceq b$.
Any two paths $\pi_1(u_c,u_d),\pi_2(u_c,u_d)$ are non-dominated (to each other) if the corresponding cost vectors do not dominate each other.
The set of all the non-dominated paths between $u_c$ and $u_d$ is called the {\it Pareto-optimal} set.
A maximal subset of the Pareto-optimal set where any two paths in this subset do not have the same cost vector is called a cost-unique Pareto-optimal set. {\blue The set of all the cost vectors of the paths in a Pareto-optimal set is called the Pareto-optimal front $\mathcal{C}^*$ ($i.e.$ a set of non-dominated cost vectors).}

In this work, we aim to compute $\mathcal{C}^*$ when the robot moves in a dynamic graph, $i.e.$ the cost vectors of edges in $\mathcal{G}$ can change.
Note that $\mathcal{C}^*$ is computed repetitively since (1) $u_c$ changes as robot moves and (2) the cost vectors of edges can change during the navigation.

	\section{Preliminaries}\label{sec:preliminary}
	
\subsection{D* Lite}\label{mopbd:sec:pre:dstar}
D* Lite~\cite{koenig2005fast} is an incremental version of A*~\cite{astar} that searches backwards from $u_d$ to $u_c$ so that the constructed search tree, which stores partial solution paths, can be reused as $u_c$ changes.
To make the presentation consistent, for the rest of the work, we present all the methods by searching \emph{backwards} from $u_d$ to $u_c$.

During the search, D* Lite maintains two types of cost-to-come at a node $u \in \mathcal{U}$: the $\gv$-value $\gv(u)$ and $\vv$-value $\vv(u)$.\footnote{We follow the convention in \cite{aine2013anytime}, where $\gv$- and $\vv$-values are introduced.} Value $\vv(u)$ stores the cost of the best path found between $u_d$ and $u$ during its last expansion, while $\gv(u)$ is computed from the $\vv$-values of $ngh(u)$, and thus, is potentially better informed than $\vv(u)$. Formally,
\begin{gather}\label{g-v-value}
	\gv(u) =
	\begin{cases}
		0 & \mbox{if } u = u_d \\
		\min_{u' \in ngh(u)} (\vv(u') + c(u,u')) & \mbox{otherwise }
	\end{cases}
\end{gather}
Based on the $\gv$- and $\vv$-values, a node $u$ is \emph{consistent} if $\vv(u)=\gv(u)$, and \emph{inconsistent} otherwise.
An inconsistent node $u$ is either underconsistent ($\vv(u) > \gv(u)$) or over-consistent ($\vv(u) < \gv(u)$).
{\blue To initialize, the $\gv$-values of all nodes but $u_d$ are set to $\infty$ and $\gv(u_d)$ is set to zero, while the $\vv$-values of all nodes are set to $\infty$.
Clearly, $u_d$ is the only inconsistent (and overconsistent as $\vv(u_d) < \gv(u_d)$) node at initialization.}

Let $h(u)$ denote the cost-to-go, which underestimates the cost of paths between $u$ and $u_c$, and define $\fv(u) := \gv(u) + \hv(u)$.
D* Lite defines the key of nodes as $key(u) = [k_1(u),k_2(u)]$ with $k_1(u) = \min\{\vv(u), \gv(u)\} + \hv(u)$ and $k_2(u) = \min\{\vv(u), \gv(u)\}$.
Let OPEN denote a priority queue containing all inconsistent nodes to be expanded, where the nodes are prioritized by comparing their keys in the lexicographic order.
In other words, $key(u) < key(w), u,w \in \mathcal{U}$ if $k_1(u) < k_1(w)$ or both $k_1(u) = k_1(w)$ and $k_2(u) < k_2(w)$.
The OPEN in D* Lite always contains all inconsistent nodes, and in each search iteration, the node with the minimum key is selected for expansion.

To expand an inconsistent node $u$, $\vv(u)$ is made equal to $\gv(u)$, which makes $u$ consistent, and for every node $w\in ngh(u)$, $\gv(w)$ is updated based on Eqn. (\ref{g-v-value}).
Additionally, a parent pointer $parent(u)$ is maintained at node $u$ when $u$ is expanded so that a path between $u_d$ and $u$ can be easily reconstructed by iteratively following the parent pointers.
D* Lite terminates when no node in OPEN has a smaller key than $key(u_c)$, which guarantees that $\vv(u_c)$ has reached the minimum and an optimal solution path between $u_d$ and $u_c$ can be reconstructed.

When computing the initial solution path $\pi(u_d,u_o)$ (note that $u_c=u_o$), D* Lite is equivalent to (backwards) A* search.
After the generation of $\pi(u_d,u_o)$, if edge costs change, D* Lite recomputes the $\gv$-values of nodes that are immediately affected by these edges.
Among these nodes, inconsistent ones are inserted into OPEN with updated keys.
Then, D* Lite runs in the same manner by expanding inconsistent nodes until all remaining nodes in OPEN have keys no less than $key(u_c)$.

\subsection{MOA*}\label{mopbd:sec:pre:moa}
The basic difference between MO-SPP and SPP is that there are multiple non-dominated partial solution paths between any pair of nodes in the graph in general.
Consequently, different from A* where $\gv,\hv,\fv$-values are computed for each node, MOA*~\cite{moastar} introduces $G,H,F$ sets:
$G(u), \forall u\in \mathcal{U}$ is a set of non-dominated cost vectors, each of which represents a non-dominated path between $u_d$ and $u$.
Similarly, $H(u)$ is a set of heuristic vectors, each of which underestimates the cost of a non-dominated path between $u$ and $u_c$.
The $F$-set is defined as $F(u):=ND\{\vec{g} + \vec{h} \,|\, \vec{g} \in G(u), \vec{h} \in H(u)\}$, where $ND(\cdot)$ is an operator that takes a set of vectors (denoted as $B$) as input and computes its non-dominated subset (denoted as $ND(B)$), $i.e.$ for any $a,b\in ND(B)$, $a$ and $b$ are non-dominated.
To simplify the presentation without losing generality, we consider the case where $H(u)$ of a node $u$ contains only a single heuristic vector $\vec{h}(u)$ that (component-wise) underestimates the cost vector of all paths between $u$ and $u_c$.

{\blue At any stage of the search process, let $\mathcal{C}$ denote the set of non-dominated cost vectors of the solutions found by the search thus far. Initially, $\mathcal{C}$ is empty. The output of the search process is $\mathcal{C^*}$ which is the true Pareto-optimal front for a given problem instance.}
In every search iteration, MOA* selects a node $u$ from OPEN so that there exists $\vec{f}\in F(u)$ that is non-dominated by any vector $\vec{f}'\in F(u')$ for any other node $u' \in$ OPEN, $u'\neq u$.
MOA* then expands the selected node $u$ by extending \emph{all} partial solutions represented by vectors in $G(u)$.
For each node $w \in ngh(u)$, a set of new partial solution paths represented by cost vectors $G' = \{\vec{g}(u) + \vec{c}(w,u) \,|\, \vec{g}(u) \in G(u)\}$ is computed and $G(w) \gets ND( G(w) \bigcup G')$, so that $G(w)$ contains all non-dominated cost vectors at $w$ after expanding $u$.
In addition, $F(w)$ is updated and node $w$ is added to OPEN for future expansion if there exists $\vec{f} \in F(w)$ that is non-dominated by cost vectors in $\mathcal{C}$.

There are two features of MOA* (node-based) that distinguish it from NAMOA* (path-based, see Sec.~\ref{mopbd:sec:pre:namoa}):
\begin{itemize}
	\item when a new non-dominated partial solution is found at node $u$, node $u$ is (re-) inserted into OPEN;
	\item when a node $u$ is selected from OPEN for expansion, all non-dominated partial solutions at $u$ are extended.
\end{itemize}
These two features show that MOA* takes a \emph{node-based} expansion strategy.
As one can expect, MOA* can lead to a lot of re-expansion of nodes as there are multiple non-dominated partial solutions at each node for a MO-SPP.
In addition, node expansion can be computationally demanding as all partial solutions at this node need to be extended.

\subsection{MOD*}\label{mopbd:sec:pre:mod}

D* Lite and MOA* can be combined as MOD* algorithm\footnote[5]{The MOD* algorithm presented in this section simplifies the method in \cite{oral2015mod} to highlight the key idea. Readers can refer to \cite{oral2015mod} for more details.} by introducing a $V$-set at each node, which resembles the $\vv$-value of a node in D* Lite, and stores the set of non-dominated cost vectors during its last expansion. Formally, it has the following relationship with the $G$-sets of neighbors.
\begin{gather}\label{MOD-G-V-set}
	G(u) = 
	\begin{cases}
		\{\vec{0}\} & \mbox{if } u = u_d \\
		ND ( \bigcup_{w \in ngh(u)} (V(w) + \vec{c}(u,w)) ) & \mbox{otherwise}
	\end{cases}
\end{gather}
Correspondingly, a node $u$ is \emph{consistent} if $G(u)=V(u)$ (two sets are exactly the same) and \emph{inconsistent} otherwise.
Similarly to D* Lite, the OPEN in MOD* contains inconsistent nodes.
MOD* iteratively selects inconsistent node $u$ from OPEN for expansion until all vectors in $F(u)$ of any inconsistent nodes $u$ in OPEN are dominated by some cost vector in $\mathcal{C}$.

When the cost vector of an edge changes, MOD* first recomputes the $G$-set of each node $u$ that are immediately affected and inserts $u$ into OPEN if $u$ is inconsistent.
Then MOD* searches in the same manner by expanding inconsistent nodes until all Pareto-optimal paths are found.
 
\subsection{NAMOA*}\label{mopbd:sec:pre:namoa}
While both MOA* and MOD* expand nodes, NAMOA*~\cite{mandow2008multiobjective}, employs a \emph{path-based} expansion to mitigate the drawbacks of the node-based expansion.

Let $s=(u,\vec{g})$ denote a \emph{state}, a tuple of a node $u$ and a cost vector $\vec{g}$, which identifies a partial solution path between $u_d$ and $u$ with cost $\vec{g}$. 
Additionally, $s$ is said to be \emph{at} node $u$.
To simplify notations, let $u(s)$ and $\vec{g}(s)$ denote the node and the cost vector contained in $s$.
For a state $s$, the $\vec{f}$-vector of $s$ is defined as $\vec{f}(s):=\vec{g}(s) + \vec{h}(u(s))$.
In NAMOA*, states (rather than nodes) are stored in OPEN as candidates.
In every search iteration, NAMOA* expands a non-dominated state $s$ in OPEN, i.e. $\vec{f}(s)$ is non-dominated by the $f$-vector of any other states in OPEN.
To expand $s$, the partial solution path represented by $s$ is extended to each neighbor $w \in ngh(u(s))$, where a new state $s' = (w,\vec{g}')$ with $\vec{g}'\gets\vec{g}+\vec{c}(u,w)$ is generated.
Cost vector $\vec{g}'$ is then compared with both the cost vectors of other partial solution paths at $w$ and the cost vectors in $\mathcal{C}$.
If $\vec{g}'$ is non-dominated, $s'$ is added to OPEN for future expansion.

As every state represents a partial solution path, expanding a state is essentially expanding a path.
This path-based strategy employed by NAMOA* avoids the large number of re-expansion of nodes as in MOA*.
In addition, expanding a path is computationally cheaper than expanding a node.

	\section{MOPBD*}\label{sec:mopbd}
	
\subsection{Algorithm Overview}

MOPBD* inherits (i) the notions of $G,H,F$-sets from MOA* (Sec.~\ref{mopbd:sec:pre:moa}), (ii) the $V$-sets from MOD* (Sec.~\ref{mopbd:sec:pre:mod}), and (iii) the concept of states from NAMOA* (Sec.~\ref{mopbd:sec:pre:namoa}).
During the search, each vector $\vec{g} \in V(u),\forall u\in \mathcal{U}$ represents a non-dominated path between $u_d$ and $u$ that has been found by the planner. $G(u)$ ``looks one step ahead'' and is computed from $V(u'),\forall u' \in ngh(u)$.
Each vector in $V(u_c)$ identifies a Pareto-optimal solution path between $u_d$ and $u_c$ and we also refer to $V(u_c)$ as $\mathcal{C}$ (the set of solution cost vectors found by the planner) for presentation purposes.
Finally, we introduce a new concept of \emph{inconsistent states}, which identifies partial solution paths that need to be expanded.
\begin{definition}[consistent state]\label{mopbd:def:inconsistent_state}
	A state $s$, with $\vec{g}(s) \in G(u(s))$, is consistent if $ \vec{g} \in V(u(s))$, and inconsistent if $\vec{g} \notin V(u(s))$.
\end{definition}

\begin{figure}[htbp]
	\centering
	\includegraphics[width=\linewidth]{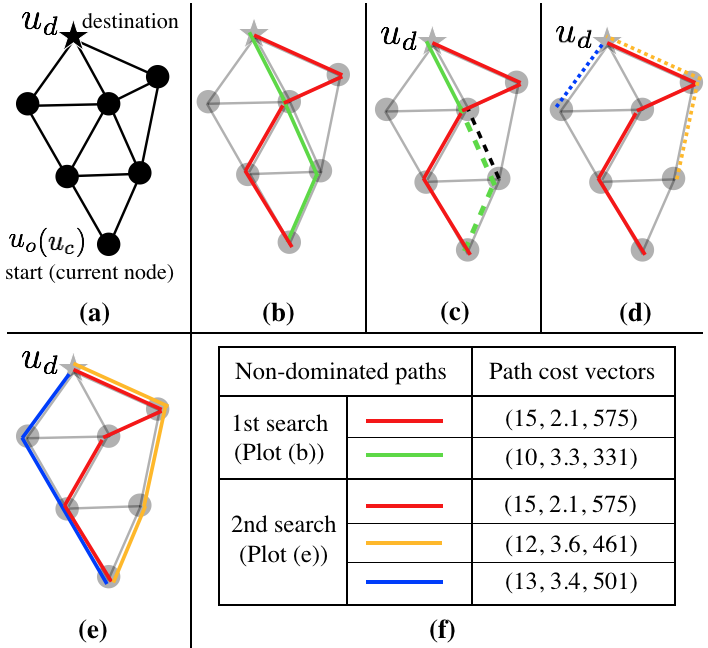}
	\vspace{-4mm}
	\caption{A conceptual visualization of MOPBD* for two planning tasks.
	Plot (a) shows the graph.
	Plot (b) shows the computed Pareto-optimal paths for the first planning task.
	Plots (c),(d) and (e) describe the second planning task.
	In Plot (c), the cost vector of the edge represented by the dashed black line is changed to an infinite vector (i.e. disconnected edge), while the green dashed line represents the portion of the paths that are affected and is thus deleted from the previous search results.
	In Plot (d), to reuse previous search efforts, MOPBD* finds new inconsistent states and adds them into OPEN for expansion.
	Plot (e) shows the computed Pareto-optimal paths in the second planning task.
	Plot (f) shows the cost vectors of the Pareto-optimal paths computed in both planning tasks.}
	\label{fig:conceptual_vis_mopbd}
	\vspace{-2mm}
\end{figure}

MOPBD* is described in Alg.~\ref{alg:all} and is conceptually visualized in Fig.~\ref{fig:conceptual_vis_mopbd}.
MOPBD* is initialized (line 1-2) by inserting a zero vector into $G(u_d)$ and creating an initial state $s_d$.
Since $V(u_d)=\emptyset$ at initialization, state $s_d$ is an inconsistent state by Def. \ref{mopbd:def:inconsistent_state} and is inserted into OPEN for expansion.
Then, MOPBD* plans paths via the \textit{ComputePath} procedure (line 3 in Alg.~\ref{alg:all}).
If the robot has not yet reached its destination, MOPBD* receives updating information about the cost vector of edges, finds all inconsistent states caused by the edge cost change (via the \textit{ProcessEdge} procedure) and re-computes paths.
If no change in edge costs, the robot navigates towards the destination along planned paths (denoted as \textit{FollowPath}). Note that for each $\vec{g} \in \mathcal{C}$, a corresponding solution path can be readily reconstructed by following the \emph{parent} pointers.



\begin{algorithm}[h]
	\small
	\caption{MOPBD*}\label{alg:all}
	\begin{algorithmic}[1]
		\State{$V(u_d)=\emptyset$, $G(u_d)$=\{$\vec{0}$\}, $s_d \gets (u_d,\vec{h}(u_d))$}
		\State{add $s_d$ to OPEN}
		\State{$\mathcal{C} \gets$ \textit{ComputePath}()}
		\While{$u_c \neq u_d$ and $\mathcal{C}\neq \emptyset$}
		\State{$\mathcal{E}' \gets $ the set of edges with updated cost vectors.}
		\If{$\mathcal{E}' \neq \emptyset$}
		\ForAll{$(w,u) \in$ $\mathcal{E'}$}
		\State{\textit{ProcessEdge}($w,u$)}
		\EndFor
		\State{$\mathcal{C} \gets$\textit{ComputePath}()}
		\Else
		\State{$u_c \gets$\textit{FollowPath}($\mathcal{C}$)}
		\EndIf
		\EndWhile
	\end{algorithmic}
\end{algorithm}
\begin{algorithm}[h!]
	\caption{\textit{ComputePath}}\label{alg:path}
	\small
	\begin{algorithmic}[1]
	    \State{$Q\gets\emptyset$}
		\While{OPEN is not empty}
		\State{$s=(u,\vec{g})$ is popped from OPEN}
		\If{$\vec{g}' \leq \vec{f}(s), \exists \vec{g}' \in V(u_c)$}
		\State{add $s$ to $Q$} \Comment{Filtered by solutions.}
		\State{\textbf{continue}} \Comment{The current iteration ends.}
		\EndIf
		\If{$\vec{g}' \leq \vec{g}(s), \exists \vec{g}' \in V(u(s))$}
		\State{remove $\vec{g}(s)$ from $G(u(s))$}
		\State{\textbf{continue}} \Comment{The current iteration ends.}
		\EndIf
		\State{\textit{UpdateVset}($s$)}
		\ForAll{$u' \in$ $ngh(u)$}
		\State{$s' \gets (u', \vec{g}(s) + \vec{c}(u',u))$}
		\If{$\vec{g}'' \leq \vec{g}(s') \; \exists \vec{g}''\in G(u(s'))$}
		\State{\textbf{continue}}
		\EndIf
		\State{add $s'$ into OPEN, add $\vec{g}(s')$ into $G(u(s'))$}
		\State{parent($s'$) $\gets s$, add $s'$ to children($s$)}
		\EndFor
		\EndWhile
		\State{OPEN$\gets Q$}
		\State{\textbf{return} $V(u_c)$}\Comment{$\mathcal{C}=V(u_c)$}
	\end{algorithmic}
\end{algorithm}

\begin{algorithm}[h!]
	\caption{\textit{Delete}($s$, $N$)}\label{alg:delete}
	\small
	\begin{algorithmic}[1]
		\ForAll{$s' \in$ children($s$)}
		\State{\textit{Delete}($s'$, $N$)}\Comment{Recursive calls.}
		\EndFor
		\State{remove $\vec{g}(s)$ from $G(u(s))$.}
		\If{$V(u(s))$ contains $\vec{g}(s)$}
		\State{remove $\vec{g}(s)$ from $V(u(s))$}
		\State{add $u(s)$ to $N$}
		\EndIf
		\State{remove parent and children pointers related to $s$}
	\end{algorithmic}
\end{algorithm}
\begin{algorithm}[h!]
	\caption{\textit{UpdateVset}($s$)}\label{alg:updateVset}
	\small
	\begin{algorithmic}[1]
		\State{$N\gets \emptyset$}
		\ForAll{$\vec{g}' \in V(u(s))$}
		\If{$\vec{g} \leq \vec{g}'$}
		\State{$s' \gets (u(s),\vec{g}')$}
    	\State{\textit{Delete}($s'$, $N$) }
		\EndIf
    	\EndFor
    	\State{Add $\vec{g}(s)$ to $V(u(s))$}
		\ForAll{$u \in N$}
		\State{$G' \gets ND(\bigcup_{u' \in ngh(u)} (V(u') + \vec{c}(u,u')) )$}
		\ForAll{$\vec{g}'$ such that $\vec{g}' \in G', \vec{g}' \notin G(u)$}
		\State{$s' \gets (u, \vec{g}')$}
		\State{add $s'$ to OPEN, add $\vec{g}$ to $G(u)$}
		\State{Update parent and children pointers related to $s'$}
		\EndFor
		\EndFor
	\end{algorithmic}
\end{algorithm}

\begin{algorithm}[h!]
	\caption{\textit{ProcessEdge}($u_1,u_2$)}\label{alg:edge}
	\small
	\begin{algorithmic}[1]
		\State{$N \gets \emptyset$ \Comment{Set of nodes where states are deleted.}  }
		\ForAll{$\vec{g} \in$ $G(u)$ with $u \in \{u_1,u_2\}$}
		\State{$s \gets (u,\vec{g})$}
		\State{$s_p \gets$ parent($s$)}
		\If{$(u(s_p),u(s))$ is the same edge as $(u_1,u_2)$}
		\State{\textit{Delete}($s$, $N$)}
		\EndIf
		\EndFor
		\ForAll{$u \in N$}
		\State{$G' \gets ND(\bigcup_{u' \in ngh(u)} (V(u') + \vec{c}(u,u')) )$}
		\ForAll{$\vec{g}'$ such that $\vec{g}' \in G', \vec{g}' \notin G(u)$}
		\State{$s' \gets (u, \vec{g}')$}
		\State{add $s'$ to OPEN, add $\vec{g}$ to $G(u)$}
		\State{Update parent and children pointers related to $s'$}
		\EndFor
		\EndFor
	\end{algorithmic}
\end{algorithm}


\subsection{Compute Pareto-optimal Paths} 

As shown in Alg. \ref{alg:path} (\textit{ComputePath}), in each search iteration, an inconsistent state $s$, with a non-dominated $\vec{f}(s)$ is popped.\footnote{In practice, the OPEN list is often implemented by prioritizing states using the lexicographic order of their $\vec{f}$-vectors~\cite{pulido2015dimensionality,ulloa2020simple}. The popped state has the lex. min. $\vec{f}$-vector in OPEN and is thus non-dominated within OPEN. We follow this practice in this work.}
Then, $\vec{f}(s)$ is first compared against every cost vector in $V(u_c)$ ($i.e.$ $\mathcal{C}$).
As $\vec{h}(u_c) = \vec{0}$, if any $\vec{g}' \in V(u_c)$ is component-wise no larger than $\vec{f}(s)$ ($i.e.$ $\vec{g}'\leq\vec{f}(s)$), then $s$ can not lead to a cost-unique Pareto-optimal solution and is thus \emph{filtered by solutions}: $s$ is inserted into another queue $Q$ and the current iteration ends.
Here, $Q$ stores all states that are filtered by solutions and are reserved for expansion in the next planning task.
This is necessary because: When the cost of an edge changes, a (previously found) solution may become no more Pareto-optimal and the states in $Q$ may lead to a Pareto-optimal solution in the next planning task.
To conceptually visualize, in Fig.~\ref{fig:conceptual_vis_mopbd} (d), the blue dashed line represents such a partial solution path, which leads to a Pareto-optimal solution after the green path is no more Pareto-optimal due to the cost change.

If not filtered by solutions, $\vec{g}(s)$ is compared against each vector $\vec{g}' \in V(u(s))$ (line 7-9 in Alg.~\ref{alg:path}).
If $\vec{g}' \leq \vec{g}(s)$ ($i.e.$ component-wise no larger than), then $s$ cannot lead to a cost-unique Pareto-optimal solution and is thus discarded. The current iteration ends.

After these comparisons, state $s$ is made consistent by adding it to $V(u(s))$ to update $V(u(s))$ via the \textit{UpdateVset} procedure, which is elaborated in the ensuing section.
Note that this includes the case when $u(s)=u_c$, where $V(u_c)$ ($i.e.$ $\mathcal{C}$) is updated. It means a new solution path with cost vector $\vec{g}(s)$ is found between $u_d$ and $u_c$.

After \textit{UpdateVset}, $s$ is expanded (line 11-16 in Alg.~\ref{alg:path}): For each $u' \in ngh(u(s))$, a path that reaches $u'$ from $u(s)$ is generated and represented by state $s'=(u', \vec{g}(s) + \vec{c}(u',u))$.
Then $\vec{g}(s')$ is compared with each vector in $G(u(s'))$:
If there exists a vector in $G(u(s'))$ that is no larger than $\vec{g}(s')$, $\vec{g}(s')$ is then discarded;
Otherwise, state $s'$ is added to OPEN for future expansion, and vector $\vec{g}(s')$ is added to $G(u(s'))$.
Additionally, the \textit{parent} of $s'$ is marked as $s$, and $s'$ is marked as a \textit{children} of $s$.
MOPBD* keeps track of ancestors and descendants of each state during the search via these {parent} and {children} pointers of generated states, which are used in the \textit{Delete} procedure (Sec.~\ref{mopbd:sec:delete}).

The search process iterates until OPEN is empty.
Then, $Q$ is assigned to OPEN for the next planning task (line 17 in Alg.~\ref{alg:path}), and $V(u_c)$ ($i.e.$ $\mathcal{C}$) is returned, which is guaranteed to be equal to $\mathcal{C}^*$, the true Pareto-optimal front of the current planning task.


\subsection{Delete States and Update $V$-sets}\label{mopbd:sec:delete}

Before describing \textit{UpdateVset}, we introduce the \textit{Delete} procedure, which is invoked by \textit{UpdateVset}.
As its name suggests, the purpose of the \textit{Delete} procedure is to remove all descendant states that have been generated during the search of a given state $s$.
First, the \textit{Delete} procedure invokes itself recursively for each \emph{children} state of $s$ (line 1-2 in Alg.~\ref{alg:delete}).
Then, vector $\vec{g}(s)$ is removed from $G(u(s))$.
If $V(u(s))$ contains $\vec{g}(s)$, then $\vec{g}(s)$ is also removed from $V(u(s))$ and node $u(s)$ is added to set $N$, which stores all the nodes whose $V$-sets have been modified during the recursive deletion.
Note that, set $N$ is passed from outside as an argument of \textit{Delete} and is modified within \textit{Delete}.
Finally, the parent and children pointers related to $s$ are removed.

Now, we explain \textit{UpdateVset}.
As shown in Alg.~\ref{alg:updateVset}, given a state $s$, to update $V(u(s))$, \textit{UpdateVset} loops over each existing vector $\vec{g}'$ in $V(u(s))$.
If $\vec{g}(s) \leq \vec{g}'$, then both the corresponding state $s' \gets (u(s),\vec{g}')$ and all descendants of $s'$ are removed by invoking the \textit{Delete} procedure.\footnote{Note that both \textit{UpdateVset} and \textit{Delete} are necessary as edge costs can become smaller at line 5 in Alg.~\ref{alg:all}, and in the next planning task, some vectors in $V(u),u\in\mathcal{U}$ are no more non-dominated.}
Additionally, let $N$ denote a set of nodes, which is initialized as an empty set (line 1 in Alg.~\ref{alg:updateVset}), and is passed to the \textit{Delete} procedure to store all the nodes whose $V$-sets are modified during \textit{Delete}.
The usage of $N$ is explained in the next paragraph.
Then, $\vec{g}(s)$ is added to $V(u(s))$.
By doing so (line 1-6 in Alg.~\ref{alg:updateVset}), $V(u), \forall u \in \mathcal{U}$ always contains the cost vectors of non-dominated paths between $u_d$ and $u$ (Invariant-1).

The second part of the \textit{UpdateVset} procedure (line 7-12 in Alg.~\ref{alg:updateVset}) seeks to find all new inconsistent states that should be generated and expanded after the modification of $V$-sets:
As aforementioned, $V(u)$ for each $u \in N$ has been modified during the \textit{Delete} procedure, which means, some vector $\vec{g}_k$ has been removed from $V(u)$.
It's possible that the vectors $\{\vec{g}_l\}$ that are previously ($i.e.$ before the removal of $\vec{g}_k$) dominated by $\vec{g}_k$ becomes non-dominated after the removal of $\vec{g}_k$, and may lead to Pareto-optimal solutions.
To find $\{\vec{g}_l\}$, \textit{UpdateVset} loops over each node $u \in N$ and computes the new $G$-set of $u$ (after \textit{Delete}) based on Eqn. (2), which is denoted as $G'$.
Then, for each vector $\vec{g}'$ that is contained in $G'$ but not contained in $G(u)$, a corresponding new state $s'\gets (u,\vec{g}')$ is generated and inserted to OPEN for future expansion.
Also, $\vec{g}'$ is added to $G(u)$, and the parent and children pointers of $s'$ are updated correspondingly.
By doing so (line 7-12 in Alg.~\ref{alg:updateVset}), all new inconsistent states at each node in $N$ are found and are added to OPEN.


\subsection{Process Edge Change}

After \textit{ComputePath}, Alg.~\ref{alg:all} either follows the planned paths or finds changes in edge costs.
As shown in Alg.~\ref{alg:edge} (\textit{ProcessEdge}), when the cost vector of an edge $(u_1,u_2)$ changes, for each $\vec{g} \in G(u)$ where $u\in\{u_1,u_2\}$, if the corresponding state $s=(u,\vec{g})$ and its parent state $parent(s)$ are at the both ends of edge $(u_1,u_2)$ (line 5 in Alg.~\ref{alg:edge}), then $s$ represents a (partial solution) path that goes through $(u_1,u_2)$.
This path is thus affected by the change of the cost vector, and both $s$ and all descendant states of $s$ are deleted by invoking the \textit{Delete} procedure (line 6).
Line 1-6 in Alg.~\ref{alg:edge} ensures that when the edge cost changes, the $V$-set of any node $u \in \mathcal{U}$ still contains the cost vectors of non-dominated paths between $u_d$ and $u$ ($i.e.$ Invariant-1).

In the meanwhile, similarly to the aforementioned \textit{UpdateVset}, the set of all nodes, whose $V$-sets are modified during the \textit{Delete} procedure, are stored in set $N$. Then, line 7-12 in Alg.~\ref{alg:edge} are the same as line 7-12 in Alg.~\ref{alg:updateVset}, which finds all new inconsistent states that should be generated after the modification of $V$-sets.

	\section{Analysis and Discussion}
	
\subsection{Pareto-optimality}
\textit{ComputePath} is invoked either at line 3 or line 9 in Alg.~\ref{alg:all}.
At both places, before entering \textit{ComputePath}, it's ensured that all inconsistent states ($i.e.$ $\{(u,\vec{g}) \,|\, \vec{g}\in G(u)\backslash V(u), \forall u \in \mathcal{U}\}$ are in OPEN.
After entering \textit{ComputePath}, in each search iteration, an inconsistent state $s$ is popped from OPEN and must either be (i) inserted into $Q$ ($i.e.$ line 4-6 in Alg.~\ref{alg:path}), (ii) discarded ($i.e.$ line 7-9 in Alg.~\ref{alg:path}), or (iii) expanded ($i.e.$ line 10-16 in Alg.~\ref{alg:path}).
For (i) and (ii), it's impossible for $\vec{g}(s)$ to be part of $\mathcal{C}^*$.
For (iii), $V(u),\forall u\in\mathcal{U}$ are updated to contain all non-dominated paths between $u_d$ and $u$ and all possible non-dominated children states of $s$ (which are inconsistent) are generated and inserted into OPEN when $s$ is expanded.
\textit{ComputePath} iterates until OPEN depletes, which guarantees that $V(u_c)$ ($i.e.$ $\mathcal{C}$) is the same as $\mathcal{C}^*$, which is summarized in the following theorem.

\begin{theorem}\label{thm:optiimality}
	When \textit{ComputePath} terminates, $\mathcal{C}=\mathcal{C}^*$.
\end{theorem}


\subsection{Runtime Analysis}
In MOPBD*, each planning task requires solving a MO-SPP, which is known to be NP-hard even with two objectives~\cite{hansen1980bicriterion}. It is also known that multi-objective search requires exponential space and time with respect to the size of the graph in the worst case~\cite{hansen1980bicriterion}.
The runtime of Alg.~\ref{alg:all} is determined by the number of states to be deleted (in \textit{ProcessEdge}) and the number of expansions (in \textit{ComputePath}).
In the worst case, Alg.~\ref{alg:all} needs to first recursively delete \emph{all} previous search results via the \textit{Delete} procedure and then start to search (from scratch).
MOPBD* is thus less efficient in comparison with naively searching from scratch, when there are lots of states to be deleted after the edge cost changes.




\subsection{MOPBD*-$\epsilon$: Approximated Pareto-optimal Front}
When there are more than two objectives, computing $\mathcal{C}^*$ becomes computationally expensive due to the enormous size of $\mathcal{C}^*$.
Correspondingly, how to fast approximate $\mathcal{C}^*$ becomes an important problem and several approximation algorithms (such as~\cite{goldin2021approximate,perny2008near}) have been developed.
In this work, we leverage $\epsilon$-dominance~\cite{perny2008near} to enable MOPBD* to approximate $\mathcal{C}^*$.

\begin{definition}[$\epsilon$-dominance]
	Given two vectors $\vec{a}$ and $\vec{b}$ of length $M$ and some $\epsilon \geq 0$, $\vec{a}$ $\epsilon$-dominates $\vec{b}$ (referred as $\vec{a} \succeq_{\epsilon} \vec{b}$) if $\vec{a}(m) \leq (1+\epsilon)\cdot \vec{b}(m)$, $\forall m \in \{1,2,\dots,M\}$.
\end{definition}
We propose MOPBD*-$\epsilon$ by replacing the $\leq$ comparison in MOPBD* with $\epsilon$-dominance. It's obvious that with a larger $\epsilon$, more partial solutions are pruned at each node during the search and the $G$- and $V$-sets at each node have a smaller size. Consequently, both \textit{ProcessEdge} and \textit{ComputePath} runs faster as there are fewer paths to be deleted or expanded when edge costs change.
As a result, MOPBD*-$\epsilon$ is able to trade off between the quality of the approximated solutions and the search efficiency, which is verified in the ensuing section.

	\section{Numerical Results}\label{sec:results}
	

\subsection{Simulation Settings}\label{sec:results:sim}

We selected (grid) maps of different categories (empty, maze, random, game) from an online data set \cite{stern2019multi} and generated a graph $G$ by making each grid four-connected.
We assigned every edge in $G$ a random integer vector of length $M$ with components randomly sampled from $[1, 10]$, where $M$ varies in the following sections.
To test a planning algorithm (referred to as ``planner'' hereafter), we implemented the following simulator (Fig.~\ref{fig:test_sim}).
For each test instance, the simulator does the following steps in order:

\noindent(Step-1) The planner computes the initial set of cost-unique Pareto-optimal paths $\Pi^*$.

\noindent(Step-2) The simulator randomly selects a path from $\Pi^*$ for the robot to follow.

\noindent(Step-3) After every $k$ ($k=7$ in our tests) moves of the robot, the simulator adds an obstacle node in front of the robot along the selected path. Adding an obstacle node means modifying the cost vector of each edge incident on that node to an infinite vector.

\noindent(Step-4) The simulator invokes the planner to re-compute cost-unique Pareto-optimal paths and goes to (Step-2).

The simulation terminates either when the robot arrives at $u_d$ (i.e. $u_c=u_d$), or when the planner returns no paths, which means the added obstacle in (Step-3) eliminates all feasible solutions.
We call (Step-1) the \emph{initial planning task} and (Step-3) the \emph{subsequent planning task}.
We set a time limit of \emph{one} minute for each planning task.
We implemented MOD*, NAMOA* and MOPBD* in Python.
All algorithms use the same heuristic: $\vec{h}(u), u\in \mathcal{U}$, which is a unit vector scaled by the Manhattan distance between $u$ and $u_c$.
Both MOD* and NAMOA* serve as baselines.

\subsection{Two Objectives Comparisons}

\begin{table}[tb]
	\centering
	\tabcolsep=0.2cm
	\renewcommand{\arraystretch}{1.16}
	\caption{Numerical results of MOPBD*, NAMOA* and MOD*. Exp. stands for the average number of expansions (either node expansion or path expansion, based on the algorithm). R.T. stands for the average runtime and Sol. stands for the average number of solutions computed. All averages are taken over all subsequent planning tasks of all test instances.}
	\begin{tabular}{ | l | l | l | l | l |  }
		\hline
		Grids & Algorithm & Exp. & R.T. & Sol.
		\\ \hline
		\multirow{2}{*}{\includegraphics[width=0.08\linewidth]{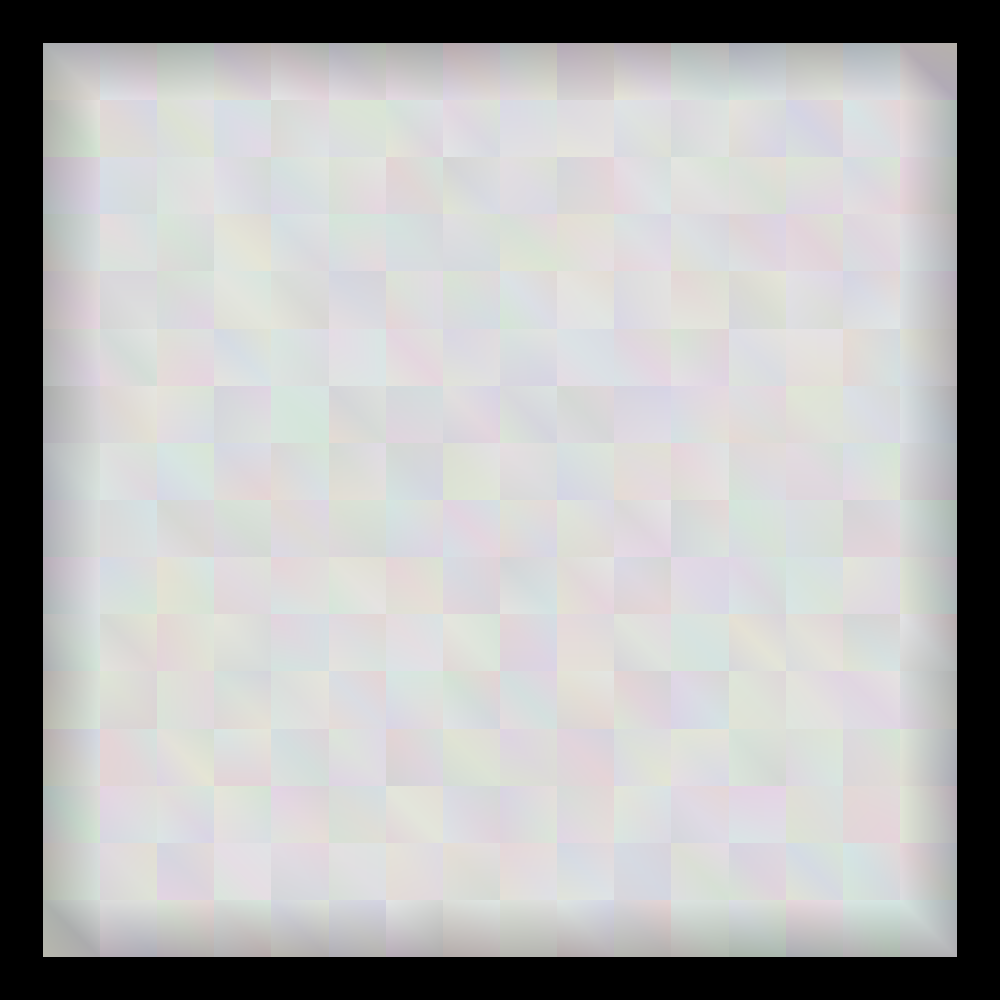}} 
		& NAMOA* & 111.8 & \textbf{0.03} & 3.0
		\\ 
		& MOD* & 39.1 & 0.35 & 3.0
		\\ 
		(16x16) & MOPBD* & \textbf{3.9} & 0.06 & 3.0
		\\ 
		\hline
		\multirow{2}{*}{\includegraphics[width=0.08\linewidth]{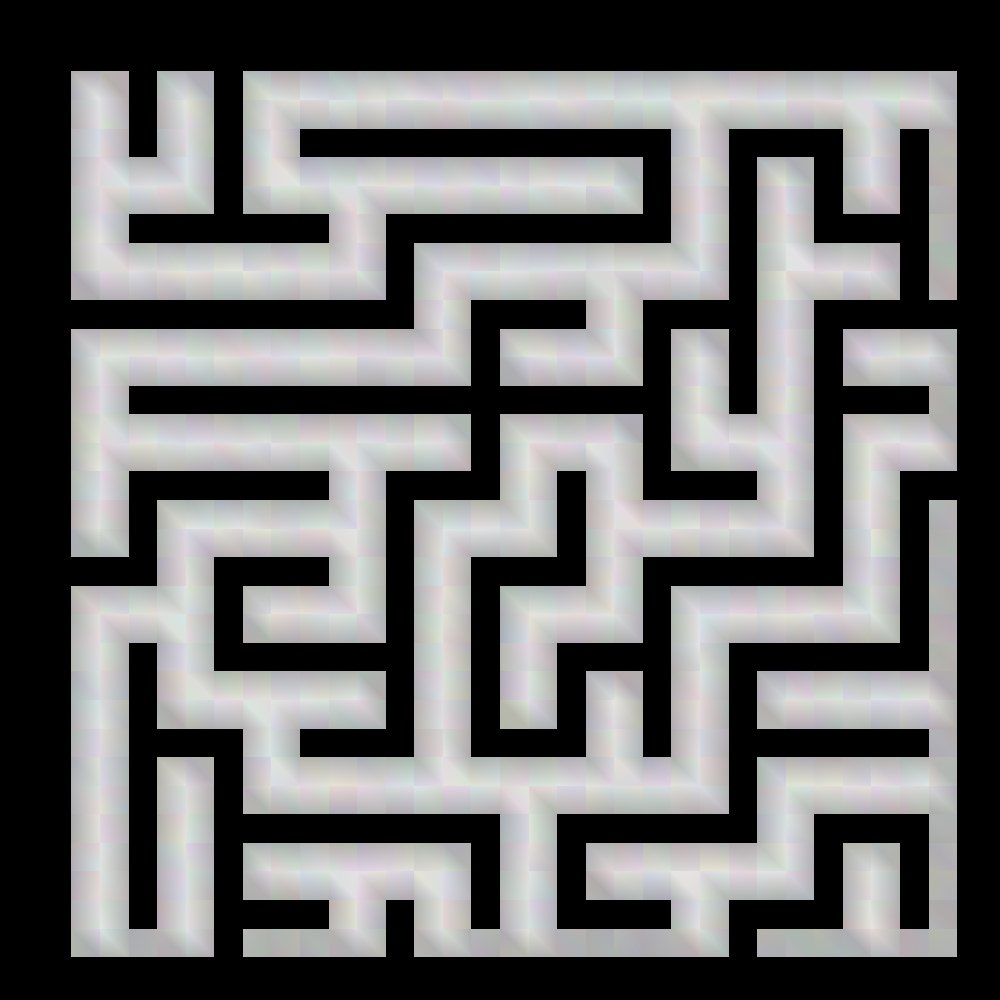}} 
		& NAMOA* & 1556.6 & 0.55 & 10.5
		\\ 
		& MOD* & 92.1 & 3.15 & 10.5
		\\ 
		(32x32) & MOPBD* & \textbf{19.7} & \textbf{0.17} & 10.5
		\\ 
		\hline
		\multirow{2}{*}{\includegraphics[width=0.08\linewidth]{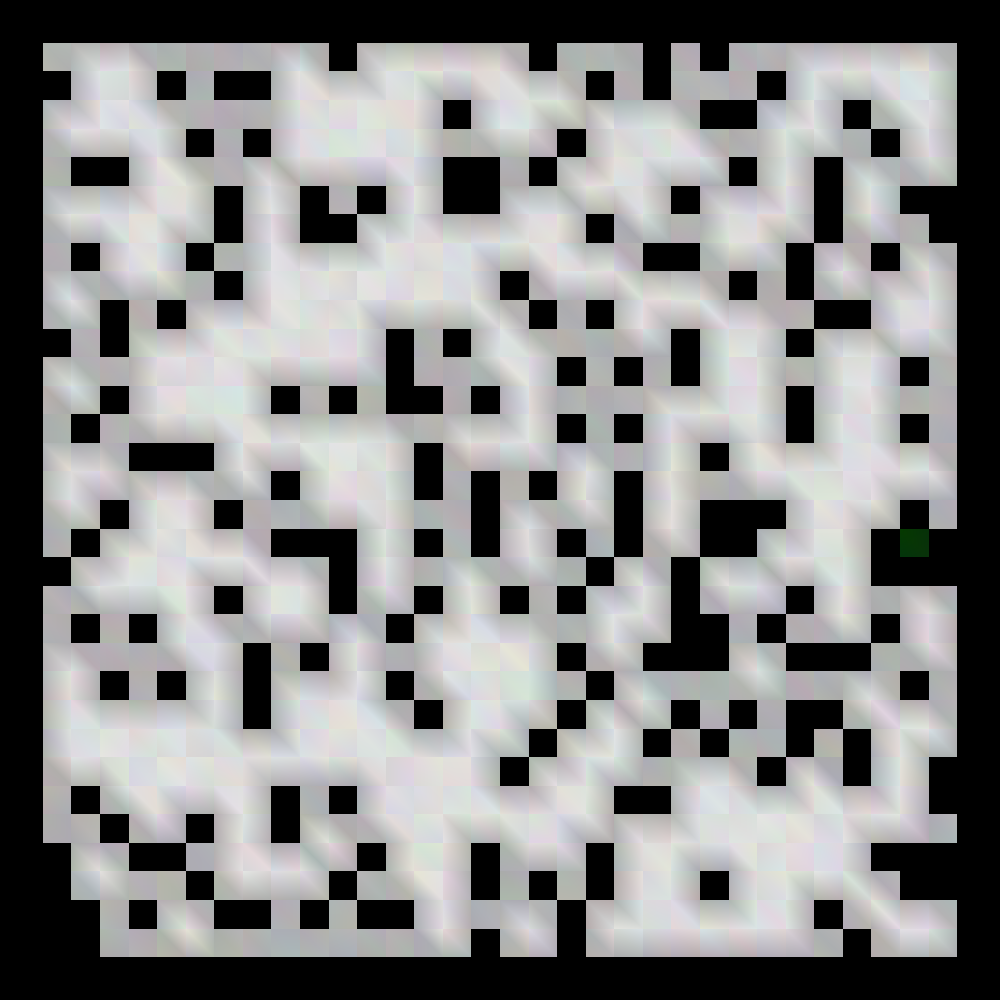}} 
		& NAMOA* & 829.5 & 0.22 & 4.9
		\\ 
		& MOD* & 311.0 & 3.51 & 4.9
		\\ 
		(32x32) & MOPBD* & \textbf{35.0} & \textbf{0.12} & 4.9
		\\ 
		\hline
		\multirow{2}{*}{\includegraphics[width=0.08\linewidth]{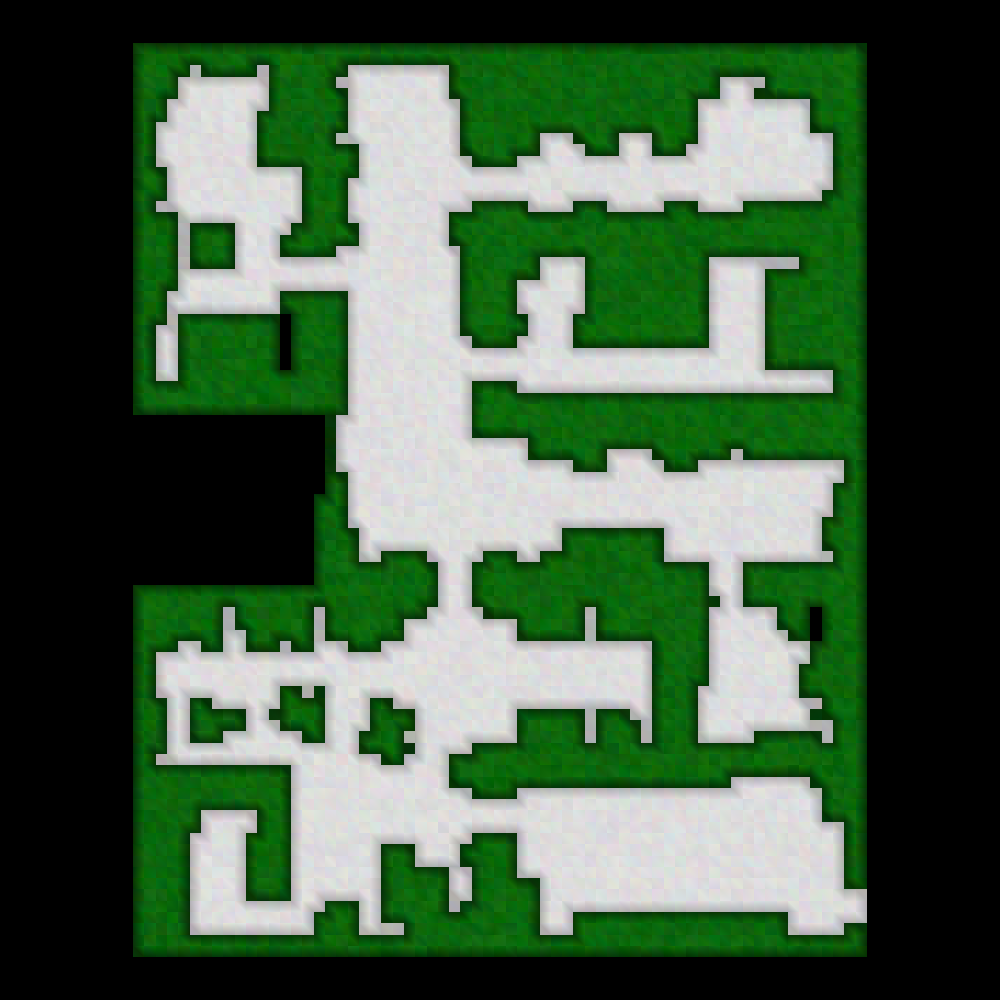}} 
		& NAMOA* & 5923.3 & 2.85 & 16.3
		\\ 
		& MOD* & 208.4 & 12.6 & 12.3
		\\ 
		(65x81) & MOPBD* & \textbf{28.0} & \textbf{2.43} & 16.3
		\\ 
		\hline
	\end{tabular}
	\label{tab:mopbd_vs_baseline}
\end{table}

We begin our tests with $M=2$.
As shown in Table~\ref{tab:mopbd_vs_baseline}, MOPBD* (path-based) runs faster than MOD* (node-based) by up to an order of magnitude.
Note that, the number of expansions cannot be directly compared between MOPBD* and MOD* as they conduct path expansion and node expansion respectively.
In the last map (a game map of size  $65\times81$), the average number of solutions found by MOD* is smaller than the other two algorithms since MOD* times out in some planning tasks.

Table~\ref{tab:mopbd_vs_baseline} also shows a comparison between NAMOA* and MOPBD*, both of which conduct path-based expansion while NAMOA* computes from scratch and MOPBD* reuses previous searches.
In terms of the number of expansions, MOPBD* outperforms NAMOA* over all maps.
In terms of runtime, MOPBD* outperforms NAMOA* in general.
However, as we observed in the $16\times16$ empty map, MOPBD* runs slower than NAMOA* on average.
The reason is that the \textit{ProcessEdge} procedure in MOPBD* is expensive when cost vectors change, as it requires deleting all affected states, and recomputing the $G$-sets of affected nodes, which is computationally demanding.

\subsection{Three Objectives and MOPBD*-$\epsilon$}\label{sec:results:mopbd_eps}

\begin{figure}[tb]
	\centering
	\includegraphics[width=1.0\linewidth]{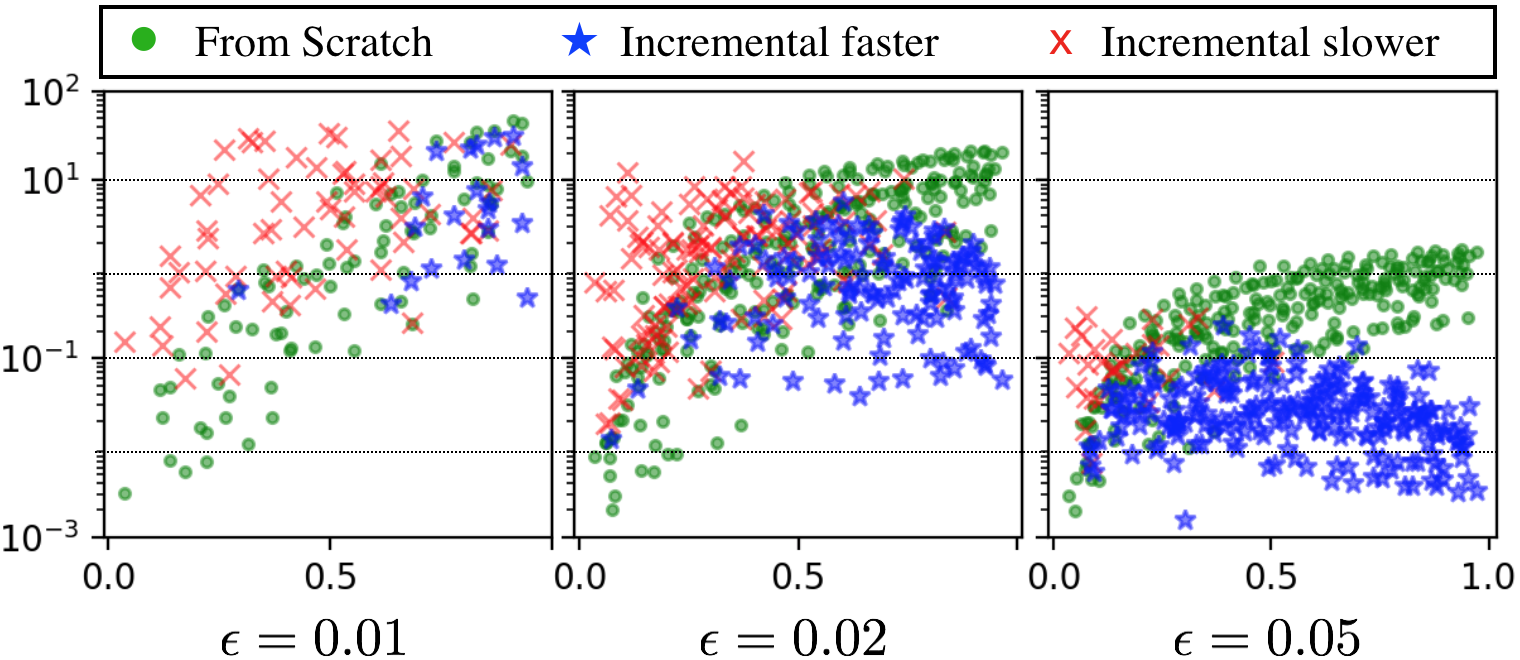}
		\vspace{-4mm}
	\caption{A detailed comparison between MOPBD*-$\epsilon$ and NAMOA*-$\epsilon$.}
	\label{fig:mopbd_eps_star}
	\vspace{-2mm}
\end{figure}

Next, we test with $M=3$ (three objectives) using the same simulator described in Sec.~\ref{sec:results:sim} to verify MOPBD*-$\epsilon$ in the aforementioned game map.
As a baseline, the dominance in NAMOA* is replaced with $\epsilon$-dominance to search from scratch for each planning task. This baseline is denoted as NAMOA*-$\epsilon$.
Here, $\epsilon$ varies among $\{0.01, 0.02, 0.05\}$. 

Given a test instance, let $l_k$ denote the average length of the Pareto-optimal paths computed in the $k$-th planning task. When $k=0$, $l_0$ denote the average length of the Pareto-optimal paths between the initial start and $u_d$.
The ratio $l_k/l_0$ estimates how far away the robot is to $u_d$ for the $k$-th planning task.
In Fig.~\ref{fig:mopbd_eps_star}, the horizontal axis represents $l_k/l_0$ within the range $[0,1]$ and the vertical axis represents the runtime of all subsequent planning tasks.

Here, green dots correspond to running NAMOA* from scratch for every planning task, and blue stars and red crosses correspond to MOPBD*.
Red crosses mean that MOPBD* is slower than NAMOA*, while the blue stars mean MOPBD* is faster.
First, both planners run faster when $\epsilon$ increases. When $\epsilon=0.01$, there are only a few data points because many planning tasks time out.
Second, as $\epsilon$ increases, there are fewer red crosses and more blue stars, which indicates that MOPBD*-$\epsilon$ gradually outperforms NAMOA*-$\epsilon$.
The reason is that when $\epsilon$ increases, only a few non-dominated partial solution paths are stored at each node, and the number of vectors in the $V$-sets (as well as the $G$-sets) at nodes become smaller, which makes the procedure \textit{ProcessEdge} computationally less demanding when edge costs change.

\subsection{Adding and Deleting Multiple Obstacles}
Finally, we test MOPBD* by adding and deleting \emph{multiple} obstacles around the robot by modifying (Step-3) of the aforementioned simulator: (Step-3) now alternates between adding and deleting two random obstacle nodes in the $5\times5$ square area centered on the robot's location.
Note that by adding an obstacle node, we modify the cost vector of each edge incident on that node to an infinite vector.
Similarly, deleting an obstacle node means assigning each edge incident on that node some finite random cost vector.
In this test, each component of the edge cost vector is randomly sampled from integers within the range $[1,5]$, and the runtime limit is set to five minutes.
We select the ``maze'' map and test with $M=2,3,4$ and $\epsilon=0$.

As shown in Table.~\ref{mopbd:tab:multiple_obst}, MOPBD* outperforms running NAMOA* from scratch for each planning task in general, based on the median and the average over succeeded cases (the better results are highlighted in the bold text).
The reason is that when randomly adding/deleting multiple obstacles around the robot, these random obstacles may affect only a few or even no paths, which allows MOPBD* to quickly fix the plan. However, as the number of objectives increases, the advantage of MOPBD* becomes less obvious and when $M=4$, NAMOA* outperforms MOPBD*. The reason is that when $M$ increases, there are more non-dominated partial solution path between a pair of nodes in general, which makes \textit{ProcessEdge} computationally expensive when edge costs change.

\begin{table}[tb]
	\centering
	\tabcolsep=0.2cm
	\renewcommand{\arraystretch}{1.16}
	\caption{Run time of planners in the format: median (average). Randomly adding or deleting obstacles near the robot in the maze map.}
	\begin{tabular}{ | l | l | l | l | }
		\hline
		$M$ & Planner & Remove Obst. & Add Obst.
		\\ \hline
		\multirow{2}{*}{2} & MOPBD* (ours) & \textbf{0.0060 (0.070)} &  \textbf{0.018 (0.12)} \\ 
		& NAMOA* & 0.042 (0.15) & 0.045 (0.19) \\ 
		\hline
		\multirow{2}{*}{3} & MOPBD* (ours) &  \textbf{0.037 (4.6)} &  \textbf{0.14 (14)} \\ 
		& NAMOA* & 0.099 (4.4) & 0.17 (6.2) \\ 
		\hline
		\multirow{2}{*}{4} & MOPBD* (ours) &  \textbf{0.062 (1.44)} & 0.24 (15) \\ 
		& NAMOA* & 0.12 (2.13) &  \textbf{0.17 (5.0)} \\ 
		\hline
	\end{tabular}
	\label{mopbd:tab:multiple_obst}
\end{table}

	\section{Conclusion and Future Work}\label{sec:conclude}
	
A new incremental multi-objective path planning algorithm MOPBD* is presented, which computes all cost-unique Pareto-optimal solutions in a dynamic graph where edge costs can change.
The numerical results verify the efficiency of MOPBD* and its variant MOPBD*-$\epsilon$ with two, three and four objectives in comparison with baseline methods.
For future work, one can consider either incorporating other multi-objective techniques \cite{ulloa2020simple,goldin2021approximate} into the algorithm to further improve performance, or leverage other incremental search techniques~\cite{aine2013anytime} to further expedite the search. One can also leverage MOPBD* to improve multi-objective multi-agent planners~\cite{ren21mocbs,ren21momstar}.
	
	\section*{Acknowledgment}
    This material is based upon work supported by the National Science Foundation under Grant No. 2120219 and 2120529. Any opinions, findings, and conclusions or recommendations expressed in this material are those of the author(s) and do not necessarily reflect the views of the National Science Foundation.
	
	\bibliographystyle{plain}
	\bibliography{ref}

\begin{thebibliography}{10}

\bibitem{aine2013anytime}
Sandip Aine and Maxim Likhachev.
\newblock Anytime truncated d*: Anytime replanning with truncation.
\newblock In {\em Sixth Annual Symposium on Combinatorial Search}, 2013.

\bibitem{bronfman2015maximin}
Andr{\'e}s Bronfman, Vladimir Marianov, Germ{\'a}n Paredes-Belmar, and Armin
  L{\"u}er-Villagra.
\newblock The maximin hazmat routing problem.
\newblock {\em European Journal of Operational Research}, 241(1):15--27, 2015.

\bibitem{deo1984shortest}
Narsingh Deo and Chi-Yin Pang.
\newblock Shortest-path algorithms: Taxonomy and annotation.
\newblock {\em Networks}, 14(2):275--323, 1984.

\bibitem{ehrgott2005multicriteria}
Matthias Ehrgott.
\newblock {\em Multicriteria optimization}, volume 491.
\newblock Springer Science \& Business Media, 2005.

\bibitem{goldin2021approximate}
Boris Goldin and Oren Salzman.
\newblock Approximate bi-criteria search by efficient representation of subsets
  of the pareto-optimal frontier.
\newblock In {\em Proceedings of the International Conference on Automated
  Planning and Scheduling}, volume~31, pages 149--158, 2021.

\bibitem{hansen1980bicriterion}
Pierre Hansen.
\newblock Bicriterion path problems.
\newblock In {\em Multiple criteria decision making theory and application},
  pages 109--127. Springer, 1980.

\bibitem{astar}
P.~E. {Hart}, N.~J. {Nilsson}, and B.~{Raphael}.
\newblock A formal basis for the heuristic determination of minimum cost paths.
\newblock {\em IEEE Transactions on Systems Science and Cybernetics},
  4(2):100--107, 1968.

\bibitem{koenig2005fast}
Sven Koenig and Maxim Likhachev.
\newblock Fast replanning for navigation in unknown terrain.
\newblock {\em IEEE Transactions on Robotics}, 21(3):354--363, 2005.

\bibitem{koenig2004lifelong}
Sven Koenig, Maxim Likhachev, and David Furcy.
\newblock Lifelong planning a*.
\newblock {\em Artificial Intelligence}, 155(1-2):93--146, 2004.

\bibitem{loui1983optimal}
Ronald~Prescott Loui.
\newblock Optimal paths in graphs with stochastic or multidimensional weights.
\newblock {\em Communications of the ACM}, 26(9):670--676, 1983.

\bibitem{mandow2008multiobjective}
Lawrence Mandow and Jos{\'e} Luis~P{\'e}rez De~La~Cruz.
\newblock Multiobjective a* search with consistent heuristics.
\newblock {\em Journal of the ACM (JACM)}, 57(5):1--25, 2008.

\bibitem{oral2015mod}
Tugcem Oral and Faruk Polat.
\newblock Mod* lite: an incremental path planning algorithm taking care of
  multiple objectives.
\newblock {\em IEEE Transactions on Cybernetics}, 46(1):245--257, 2015.

\bibitem{perny2008near}
Patrice Perny and Olivier Spanjaard.
\newblock Near admissible algorithms for multiobjective search.
\newblock In {\em 18th European Conference on Artificial Intelligence ECAI-08},
  volume 178, pages 490--494. IOS Press, 2008.

\bibitem{pulido2015dimensionality}
Francisco-Javier Pulido, Lawrence Mandow, and Jos{\'e}-Luis P{\'e}rez-de-la
  Cruz.
\newblock Dimensionality reduction in multiobjective shortest path search.
\newblock {\em Computers \& Operations Research}, 64:60--70, 2015.

\bibitem{ren21mocbs}
Zhongqiang Ren, Sivakumar Rathinam, and Howie Choset.
\newblock Multi-objective conflict-based search for multi-agent path finding.
\newblock In {\em 2021 IEEE International Conference on Robotics and Automation
  (ICRA)}, pages 8786--8791, 2021.

\bibitem{ren21momstar}
Zhongqiang Ren, Sivakumar Rathinam, and Howie Choset.
\newblock Subdimensional expansion for multi-objective multi-agent path
  finding.
\newblock {\em IEEE Robotics and Automation Letters}, 6(4):7153--7160, 2021.

\bibitem{roijers2013survey}
Diederik~M Roijers, Peter Vamplew, Shimon Whiteson, and Richard Dazeley.
\newblock A survey of multi-objective sequential decision-making.
\newblock {\em Journal of Artificial Intelligence Research}, 48:67--113, 2013.

\bibitem{shan2020receding}
Tixiao Shan, Wei Wang, Brendan Englot, Carlo Ratti, and Daniela Rus.
\newblock A receding horizon multi-objective planner for autonomous surface
  vehicles in urban waterways.
\newblock In {\em 2020 59th IEEE Conference on Decision and Control (CDC)},
  pages 4085--4092. IEEE, 2020.

\bibitem{stentz1995focussed}
Anthony Stentz.
\newblock The focussed d* algorithm for real-time replanning.
\newblock In {\em Proceedings of the 14th International Joint Conference on
  Artificial Intelligence}, volume~95, pages 1652--1659, 1995.

\bibitem{stern2019multi}
Roni Stern, Nathan Sturtevant, Ariel Felner, Sven Koenig, Hang Ma, Thayne
  Walker, Jiaoyang Li, Dor Atzmon, Liron Cohen, TK~Kumar, et~al.
\newblock Multi-agent pathfinding: Definitions, variants, and benchmarks.
\newblock In {\em Symposium on Combinatorial Search}, page 151–158, 2019.

\bibitem{moastar}
Bradley~S. Stewart and Chelsea~C. White.
\newblock Multiobjective a*.
\newblock {\em Journal of the ACM (JACM)}, 38(4):775--814, October 1991.

\bibitem{ulloa2020simple}
Carlos~Hern{\'a}ndez Ulloa, William Yeoh, Jorge~A Baier, Han Zhang, Luis Suazo,
  and Sven Koenig.
\newblock A simple and fast bi-objective search algorithm.
\newblock In {\em Proceedings of the International Conference on Automated
  Planning and Scheduling}, volume~30, pages 143--151, 2020.

\bibitem{urmson2008autonomous}
Chris Urmson, Joshua Anhalt, Drew Bagnell, Christopher Baker, Robert Bittner,
  MN~Clark, John Dolan, Dave Duggins, Tugrul Galatali, Chris Geyer, et~al.
\newblock Autonomous driving in urban environments: Boss and the urban
  challenge.
\newblock {\em Journal of Field Robotics}, 25(8):425--466, 2008.

\bibitem{wurman2008coordinating}
Peter~R Wurman, Raffaello D'Andrea, and Mick Mountz.
\newblock Coordinating hundreds of cooperative, autonomous vehicles in
  warehouses.
\newblock {\em AI magazine}, 29(1):9--9, 2008.

\end{thebibliography}

	
\end{document}